\documentclass[11pt]{article}

\usepackage{amssymb,amsfonts,amsmath,amsthm,bm}
\usepackage{algorithm2e,algorithmic,framed,url}
\usepackage{graphicx}
\usepackage{framed}
\usepackage{url}
\usepackage{amssymb}
\usepackage{amsfonts}
\usepackage{amsmath}
\usepackage{amsfonts}
\usepackage{graphicx}
\usepackage{subfigure}
\usepackage{enumerate}
\usepackage{bm}
\usepackage{amsmath} 
\usepackage{amsthm}
\usepackage{wrapfig}
\usepackage{array}
\usepackage{color}
\usepackage{graphics}
\usepackage{multirow}

\newcommand{\FNorm }[1]{\mbox{}\|#1\|_\mathrm{F}  }
\newcommand{\FNormS}[1]{\mbox{}\|#1\|_\mathrm{F}^2}

\newcommand{\TNorm }[1]{\mbox{}\|#1\|_2  }
\newcommand{\TNormS}[1]{\mbox{}\|#1\|_2^2}

\newtheorem{theorem}{\bf Theorem}[]
\newtheorem{lemma}[theorem]{Lemma}
\newtheorem{definition}[theorem]{Definition}

\newcommand{\transp}{^{\textsc{T}}}

\newcommand{\mat}[1]{{\ensuremath{\bm{\mathrm{#1}}}}}

\def\rank{\hbox{\rm rank}}

\def\b{{\mathbf b}}

\def\matA{\mat{A}}
\def\matB{\mat{B}}

\def\matD{\mat{D}}
\def\matE{\mat{E}}

\def\matI{\mat{I}}
\def\matL{\mat{L}}

\def\matP{\mat{P}}
\def\matQ{\mat{Q}}

\def\matS{\mat{S}}

\def\matU{\mat{U}}
\def\matV{\mat{V}}
\def\matW{\mat{W}}
\def\matX{\mat{X}}
\def\matY{\mat{Y}}
\def\matZ{\mat{Z}}
\def\matSig{\mat{\Sigma}}

\def\matOmega{\mat{\Omega}}

\DeclareMathSymbol{\Prob}{\mathbin}{AMSb}{"50}
\newcommand\remove[1]{}

\def\math#1{$#1$}

\def\frac#1#2{{#1\over #2}}

\def\eqan#1{\begin{eqnarray*}
#1
\end{eqnarray*}}

\DeclareMathSymbol{\R}{\mathbin}{AMSb}{"52}

\newcommand{\argmin}{\operatorname*{argmin}}

\def\x{{\mathbf x}}
\def\y{{\mathbf y}}

\def\a{{\mathbf a}}
\def\b{{\mathbf b}}

\def\dotfil{\leaders\hbox to 1.5mm{.}\hfill}
\newcounter{rmnum}
\def\RN#1{\setcounter{rmnum}{#1}\uppercase\expandafter{\romannumeral\value{rmnum}}}
\def\rn#1{\setcounter{rmnum}{#1}\expandafter{\romannumeral\value{rmnum}}}

\newcommand{\cut}{\operatorname{cut}}
\newcommand{\Ncut}{\operatorname{Ncut}}
\newcommand{\assoc}{\operatorname{assoc}}
\def\R{\mathbb{R}}

\graphicspath{{../Images/}}

\usepackage{graphicx}
\usepackage{amsmath,amsthm,amssymb,bm}
\usepackage{subfigure}

\def\transp{T}

\theoremstyle{remark}

\begin{document}

\title{Spectral Clustering via the Power Method -- Provably}

\author{
Christos Boutsidis\\Yahoo! \\ New York, New York \\ boutsidis@yahoo-inc.com
\and
Alex Gittens\\International Computer Science Institute \\ Berkeley, California \\ gittens@icsi.berkeley.edu
\and
Prabhanjan Kambadur\\Bloomberg L.P. \\ New York, New York \\ Pkambadur@bloomberg.net
}

\date{}
\maketitle

\begin{abstract}
Spectral clustering is one of the most important algorithms in data
mining and machine intelligence; however, its computational complexity 
limits its application to truly large scale data analysis. 
The computational bottleneck in spectral clustering is computing a few of the
top eigenvectors of the (normalized) Laplacian matrix corresponding to the
graph representing the data to be clustered. 
One way to speed up the computation of these eigenvectors is to use the ``power
method'' from the numerical linear algebra literature. 
Although the power method has been empirically used to speed up spectral
clustering, the theory behind this approach, to the best of our knowledge,
remains unexplored. 
This paper provides the \emph{first} such rigorous theoretical justification,
arguing that a small number of power iterations suffices to obtain near-optimal 
partitionings using the approximate eigenvectors. 
Specifically, we prove that solving the $k$-means clustering problem on the
approximate eigenvectors obtained via the power method gives an additive-error
approximation to solving the $k$-means problem on the optimal eigenvectors. 
\end{abstract}

\section{Introduction}
\label{sec:introduction}

\begin{figure}[t]
\centering
\includegraphics[width=0.425\textwidth] {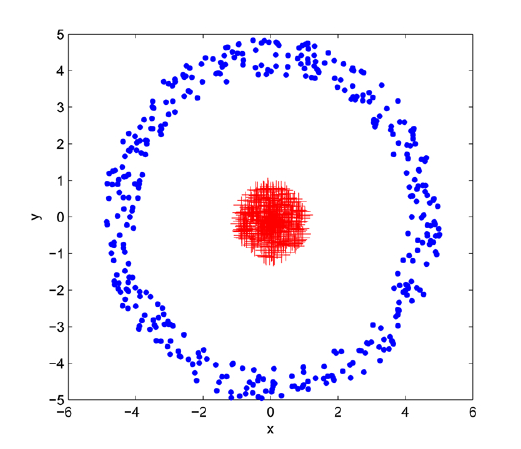} 
\caption{$2$-D data amenable to spectral clustering.} 
\label{fig:non-separable}
\end{figure}
\vspace{-5pt}
Consider clustering the points in Figure~\ref{fig:non-separable}.
The data in this space are non-separable and there is no apparent clustering
metric which can be used to recover this clustering structure.
In particular, the two clusters have the same centers (centroids); hence,
distance-based clustering methods such as
$k$-means~\cite{ostrovsky2006effectiveness} will fail.
Motivated by such shortcomings of traditional clustering approaches,
researchers have produced a body of more flexible and data-adaptive clustering
approaches, now known under the umbrella of \emph{spectral clustering}. 
The crux of these approaches is to model the points to be clustered as vertices
of a graph, where weights on edges connecting the vertices are assigned
according to some similarity measure between the points.
Next, a new, hopefully separable, representation of the points is formed by
using the eigenvectors of the (normalized) Laplacian matrix associated with
this similarity graph. 
This new, typically low-dimensional, representation is often called ``spectral
embedding'' of the points.
We refer the reader to~\cite{Fie73,Ulr07,SM00} for the foundations of spectral
clustering and to ~\cite{BP01,NG02,LZ04,MP04,SW05} for applications in data
mining and machine learning. 
We explain spectral clustering and the baseline algorithm in detail in
Section~\ref{sec:problem}.

The computational bottleneck in spectral clustering is the computation of the
eigenvectors of the Laplacian matrix. 
Motivated by the need for faster algorithms to compute these eigenvectors,
several techniques have been developed in order to speedup this
computation~\cite{
ST09,
yan2009fast, 
fowlkes2004spectral,
pavan2005efficient, 
bezdek2006approximate, 
wang2009approximate,
nystrom1930praktische, 
baker1977numerical}. 
Perhaps the most popular of the above mentioned techniques is the ``power
method''~\cite{lin2010power}. 
The convergence of the power method is theoretically well understood when it
comes to measure the principal angle between the space spanned by the true and
the approximate eigenvectors~(see Theorem 8.2.4 in~\cite{golub2012matrix}).  
We refer readers to~\cite{Woo14} for a rigorous theoretical analysis of the use
of the power method for the low-rank matrix approximation problem.  
However, these results do not imply that the approximate eigenvectors of the
power method are useful for spectral clustering.  

\paragraph{Contributions.}
In this paper, we  argue that the eigenvectors computed via the power
method are useful for spectral clustering, and that the loss in clustering
accuracy is small. 
We prove that solving the $k$-means problem on the approximate
eigenvectors obtained via the power method gives an additive-error
approximation to solving the $k$-means problem on the optimal eigenvectors
(see Lemma~\ref{thm:main} and Thm~\ref{thm:MAIN}).

\section{Background}
\label{sec:clustering}
\subsection{Spectral Clustering }
\label{sec:problem}
We first review one mathematical formulation of spectral clustering.
Let $\x_1, \x_2, \ldots, \x_n \in \R^d$ be $n$ points in $d$ dimensions.
The goal of clustering is to partition these points into $k$ disjoint sets,
for some given $k$.
To this end, define a weighted undirected graph $G(V,E)$ with $|V| = n$ nodes
and $|E|$ edges:
each node in $G$ corresponds to an $\x_i$; the weight of each edge encodes
the similarity between its end points.
Let $\matW \in \R^{n \times n}$ be the similarity matrix ---
$\matW_{ij}=e^{-(\|\x_i-\x_j\|^2)/\sigma}, i\ne{}j$ and  $\matW_{ii}=0$ ---
that gives the similarity between $\x_i$ and $\x_j$. 
Here, $\sigma$ is a tuning parameter.
Given this setup, spectral clustering for $k=2$ corresponds to the following
graph partitioning problem:
\begin{definition}[The Spectral Clustering Problem for $k=2$~\cite{SM00}]
\label{def1}
Let $\x_1, \x_2, \ldots, \x_n \in \R^{d}$ and $k=2$ be given.
Construct graph $G(V,E)$ as described in the text above.
Find subgraphs $A$ and $B$ of $G$ that minimize the following:
\begin{equation*}
\Ncut(A, B) = 
\cut(A, B) \cdot \left(
\frac{1}{\assoc(A, V)} + \frac{1}{\assoc(B, V)}
\right),
\label{eq:clustering_ncut}
\end{equation*}
\textnormal{where,}
$\cut(A, B) = \sum_{\substack{\x_i \in A, \x_j \in B}}
\matW_{ij};$
$\assoc(A, V) = \sum_{\substack{\x_i \in A, \x_j \in V}}
\matW_{ij};$
$\quad \assoc(B, V) = \sum_{\substack{\x_i \in B, \x_j \in V}}
\matW_{ij}$.
\end{definition}

This definition generalizes to any $k > 2$ in a straightforward manner (we omit
the details).
Minimizing $\Ncut(A,B)$ in a weighted undirected graph is an NP-Complete
problem (see appendix in~\cite{SM00} for proof).
Motivated by this hardness result, Shi and Malik~\cite{SM00} suggested a
relaxation to this problem that is tractable in polynomial time using the
Singular Value Decomposition (SVD).
First, \cite{SM00} shows that for \emph{any} $G, A, B$ and partition vector $\y
\in \R^{n}$ with $+1$ to the entries corresponding to $A$ and $-1$ to the
entries corresponding to $B$ the following identity holds:
$4 \cdot \Ncut(A, B) =  \y^\transp(\matD - \matW)\y / (\y^\transp\matD\y).$
Here, $\matD \in \R^{n \times n}$ is the diagonal matrix of degree nodes:
$\matD_{ii}=\sum_j \matW_{ij}.$ 
%
Hence, the spectral clustering problem in Definition~\ref{def1} can be restated
as finding such an optimum partition vector $\y$, which, as we mentioned above,
is an intractable problem. 
The real relaxation for spectral clustering asks for a real-valued vector $\y
\in \R^n$:
\begin{definition}[The real relaxation for the spectral clustering
problem for $k=2$~\cite{SM00}]
\label{def2}
Given graph $G$ with $n$ nodes, adjacency matrix $\matW,$ and degrees matrix
$\matD$ find $\y \in \R^n$ such that:
\begin{equation*}
\y = \argmin_{\y \in \R^n, \y^\transp \matD{\bf 1}_n}
      \frac{\left(\y^\transp(\matD-\matW)\y\right)}
           {\left(\y^\transp \matD \y \right)}.
\label{eq:relaxed_clustering_ncut}
\end{equation*}
\end{definition}
Once such a $\y$ is found, one can partition the graph into two subgraphs by
looking at the signs of the elements in $\y$.
When $k > 2$, one can compute $k$ eigenvectors and then apply $k$-means
clustering on the rows of a matrix, denoted as $\matY$, containing those
eigenvectors in its columns.

Motivated by these observations, Ng et.~al~\cite{NG02}~(see
also~\cite{weiss1999segmentation}) suggested the following algorithm for
spectral clustering \footnote{Precisely, Ng et.~al~suggested an additional
normalization step on $\matY$ before applying $k$-means, i.e., normalize
$\matY$ to unit row norms, but we ignore this step for simplicity.} (inputs to
the algorithm are the points $\x_1,\dots,\x_n \in \R^{d}$ and the number of
clusters $k$).
\footnotesize
\begin{enumerate}
\item Construct the similarity matrix $\matW \in \R^{n \times n}$ as
$\matW_{ij}=e^{-(\|\x_i-\x_j\|^2)/\sigma}$ (for $i \ne j$); $\matW_{ii}=0$ 
and $\sigma$ is given.
\item Construct $\matD \in \R^{n \times n}$ as the diagonal matrix of degrees
of the nodes: $\matD_{ii}=\sum_j \matW_{ij}.$
\item Construct $\tilde{\matW} = \matD^{-\frac{1}{2}}\matW\matD^{-\frac{1}{2}}
\in \R^{n \times n}$.~\footnote{Here, $\matL = \matD - \matW$ is the Laplacian
matrix of $G$ and $\tilde\matL = \matI_n - \tilde\matW$ is the so called
normalized Laplacian matrix.}
\item Find the largest $k$ eigenvectors of $\tilde{\matW}$ and assign them as
columns to a matrix $\matY \in \R^{n \times k}$
\footnote{The top $k$ eigenvectors of
$\matD^{-\frac{1}{2}}\matW\matD^{-\frac{1}{2}}$ correspond to the bottom $k$
eigenvectors of $\matI_n - \matD^{-\frac{1}{2}}\matW\matD^{-\frac{1}{2}}$.
}.
%
\item Apply $k$-means clustering on the rows of $\matY$, and use this
clustering to cluster the original points accordingly.
\end{enumerate}
\normalsize
%
%
This algorithm serves as our baseline for an ``exact spectral clustering
algorithm''.  
One way to speedup 
\footnote{This can be implemented in $O(n^2kp + k^2n)$ time,
as we need $O(n^2kp)$ time to implement all the matrix-matrix multiplications
(right-to-left) and another $O(k^2n)$ time to find $\tilde{\matY}$.
As we discuss  below, selecting $p \approx O(\ln(kn))$ and assuming that the
multiplicative spectral gap of between the $k^{th}$ and $(k+1)^{th}$ eigenvalue
of $\tilde{\matW}$ is large suffices to get very accurate clusterings.
This leads to an  $O(k n^2 \ln(kn))$ runtime for this step.} 
this baseline algorithm is to use the power method~\cite{lin2010power}
in Step 4 to quickly approximate the eigenvectors in $\matY$; that is,
\footnotesize
\begin{itemize}
\item
\emph{\bf Power method:} Initialize $\matS \in \R^{n \times k}$ with $i.i.d$
random Gaussian variables. 
Let $\tilde{\matY} \in \R^{n \times k}$ contain the left singular vectors of
the matrix 
$$\matB = (\tilde{\matW} \tilde{\matW}^\transp)^p \tilde{\matW} \matS
= \tilde\matW^{2p + 1} \matS,
$$ for some integer $p \ge 0$.
Now, use $\tilde{\matY}$ instead of $\matY$ in step 5 above.
\end{itemize}
\normalsize
The use of the power method to speedup eigenvector computation is not new. 
Power method is a classical technique in the numerical linear algebra
literature (see Section 8.2.4 in~\cite{golub2012matrix}). 
Existing theoretical analysis provides sharp bounds for the error $\TNorm{\matY
\matY^\transp - \tilde{\matY} \tilde{\matY}^\transp}$ (see Theorem 8.2.4
in~\cite{golub2012matrix}; this theorem assumes that $\matS$ has orthonormal
columns and it is not perpendicular to $\matY$.)
%
\cite{halko2011finding,Woo14} also used the power method with random Gaussian
initialization and applied it to the low-rank matrix approximation problem.
The approximation bounds proved in those papers are for the term
$\TNorm{\tilde\matW - \tilde{\matY} \tilde{\matY}^\transp \tilde\matW}$. 
To the best of our knowledge, none of these results indicates that the
approximate eigenvectors are useful for spectral clustering purposes. 
%
 
\subsection{Connection to $k$-means} 
The previous algorithm indicates that spectral clustering turns out to be a
$k$-means clustering problem on the rows of $\matY$, the matrix containing the
bottom eigenvectors of the normalized Laplacian matrix. 
%
The main result of our paper is to prove that solving the $k$-means problem on
$\tilde\matY$ and using this to cluster the rows in $\matY$ gives a clustering
which is as good as the clustering by solving the $k$-means problem on $\matY$. 
%
To this end, we need some background on $k$-means clustering; we present a
linear algebraic view below. 

For $i=1:n,$ let $\y_i \in \R^k$ be a row of $\matY$ as a column vector. 
Hence, $$\matY= \left[\begin{matrix}\text{--- }\y_1^\transp\text{
    ---}\\\text{--- } \y_2^\transp\text{ ---}\\ \vdots\\ \text{---
    }\y_n^\transp\text{ ---} \end{matrix}\right] \in \R^{n \times k}.$$
Let $k$ be the number of clusters. One can define a partition of the rows of
$\matY$ by a cluster indicator matrix $\matX \in \R^{n \times k}$.
Each column \math{j=1,\ldots,k} of \math{\matX} represents a cluster.
Each row \math{i=1,\ldots,n} indicates the cluster membership of \math{\y_i}. 
So, $\matX_{ij}=1/\sqrt{s_j},$ if and only if the data point \math{\y_i} is in
the $j$th cluster ($s_j =  ||\matX^{(j)}||_{0}$; $\matX^{(j)}$ is the $j$th
column of $\matX$ and $||\matX^{(j)}||_{0}$ denotes the number of non-zero
elements of $\matX^{(j)}$).
%
%
We formally define the $k$-means problem as follows: 
\begin{definition}
\textsc{[The $k$-means clustering problem]}
Given $\matY \in \mathbb{R}^{n \times k}$ (representing $n$ data points -- rows
-- described with respect to $k$ features -- columns) and a positive integer
$k$ denoting the number of clusters, find the indicator matrix
$\matX_{\mathrm{opt}} \in \R^{n \times k}$ which satisfies,
$$
\matX_{\mathrm{opt}} = \argmin_{\matX \in \cal{X}} \FNormS{\matY - \matX
\matX^\transp \matY}.
$$
Here, $\cal{X}$ denotes the set of all $m \times k$ indicator matrices $\matX$.
Also, we will denote
$$
\FNormS{\matY - \matX_{\mathrm{opt}} \matX_{\mathrm{opt}}^\transp\matY} 
:= \mathrm{F}_{\mathrm{opt}}. 
$$
\end{definition}
This definition is equivalent to the more traditional definition involving sum
of squared distances of points from cluster
centers~\cite{boutsidis2013deterministic,cohen2014dimensionality} (we omit the
details). 
%
%
Next, we formalize the notion of a ``$k$-means approximation algorithm''. 
\begin{definition} \label{def:approx}
\textsc{[$k$-means approximation algorithm]} An algorithm is called a
``$\gamma$-approximation'' for the $k$-means clustering problem ($\gamma \geq
1$) if it takes inputs the dataset $\matY \in \R^{n \times k}$ and the number
of clusters $k$, and returns an indicator matrix $\matX_{\gamma} \in \R^{n
\times k}$ such that w.p. $1 - \delta_{\gamma}$, \eqan{ \FNormS{\matY -
\matX_{\gamma} \matX_{\gamma}^\transp \matY} \leq \gamma \min_{\matX \in
\cal{X}} \FNormS{\matY - \matX \matX^\transp \matY} = \gamma \cdot
\mathrm{F}_{\mathrm{opt}}.  }
\end{definition}

An example of such an approximation algorithm  is in~\cite{KSS04} with $\gamma
= 1+\varepsilon$ ($0 < \varepsilon < 1$) and $\delta_{\gamma}$ a constant in
$(0,1)$.
The corresponding running time is $O(n k \cdot 2^{(k/\varepsilon)^{O(1)}})$.
A trivial algorithm with $\gamma = 1$ but running time $\Omega(n^k)$ is to
try all possible  $k$-clusterings of the rows of $\matY$ and keep the best. 
  
\section{Main result}
\label{sec:theory}
Next, we argue that applying a $k$-means approximation algorithm on $\matY$ and $\tilde{\matY}$
gives approximately the same clustering results for a sufficiently large
number of power iterations $p$. 
Hence, the eigenvectors found by the power method do not sacrifice the
accuracy of the exact spectral clustering algorithm.  
This is formally shown in Theorem~\ref{thm:MAIN}. However, the key notion of approximation
between the exact and the approximate eigenvectors is captured in Lemma~\ref{thm:main}:
\begin{lemma}\label{thm:main} [See Section~\ref{sec:proof} for proof]
For any $\varepsilon, \delta > 0,$ let
\begin{equation*}
p \geq  \frac{ \frac{1}{2} \ln\left( 4 \cdot  n \cdot  \varepsilon^{-1} \cdot \delta^{-1} \cdot \sqrt{k} \right)} {
\ln\left( \gamma_k \right)},
%
\end{equation*}
where
$$
\gamma_k = \frac{ \sigma_k\left(\tilde{\matW}\right)}{\sigma_{k+1} \left(
\tilde{\matW} \right)},
$$ 
is the multiplicative eigen-gap between the $k$-th and the $(k+1)$-th singular value of $\tilde\matW$. 
Then, with probability at least
$1 - e^{-2n} - 2.35\delta$: 
$ \FNormS{\matY \matY^\transp - \tilde{\matY}\tilde\matY^\transp}
\le \varepsilon^2.
$
\end{lemma}
In words, for a sufficiently large value of $p$, 
the orthogonal projection operators on $span(\matY)$ and $span(\tilde\matY)$
are bounded, in Frobenius norm, for an arbitrarily small  $\varepsilon$. 
Here and throughout the paper we reserve $\sigma_1(\tilde\matW)$ 
to denote the largest singular value of $\tilde\matW$: 
$$\sigma_1(\tilde\matW) \ge \sigma_2(\tilde\matW) \ge \dots \ge \sigma_n(\tilde\matW) \ge 0;$$ 
and similarly for $\tilde\matL$: 
$$
\lambda_1(\tilde\matL) \ge  \lambda_2(\tilde\matL) \ge \dots \ge  \lambda_n(\tilde\matL).
$$

Next, we present our
main theorem~(see Section~\ref{sec:proofmain} for the proof). 
\begin{theorem}\label{thm:MAIN}
Construct $\tilde\matY$ via the power method with 
$$ p \ge  \frac{ \frac{1}{2} \cdot \ln\left(4 \cdot n \cdot \varepsilon^{-1} \cdot \delta^{-1} \cdot \sqrt{k} \right)}
{\ln\left(\gamma_k \right) }, $$
where 
$$
\gamma_k = \frac{ \sigma_k\left(\tilde{\matW}\right)}{\sigma_{k+1} \left(
\tilde{\matW} \right)} = \frac{1 - \sigma_{n-k+1}(\tilde\matL)}{ 1- \sigma_{n-k}(\matL)}.
$$ 
Here, $\tilde\matL = \matI_n - \tilde\matW$ is the normalized Laplacian matrix. 
Consider running on the rows of $\tilde\matY$ a $\gamma$-approximation $k$-means algorithm with failure probability $\delta_{\gamma}$.
Let the outcome be a clustering indicator matrix $\matX_{\tilde{\gamma}} \in \R^{n \times k}$. 
Also, let $\matX_{\mathrm{opt}}$ be the optimal clustering indicator matrix for
$\matY$. Then, with probability at least $1 - e^{-2n} - 2.35\delta -
\delta_{\gamma},$
$$
\FNormS{\matY - \matX_{\tilde{\gamma}} \matX_{\tilde{\gamma}}^\transp  \matY}
\leq  (1+4\varepsilon) \cdot  \gamma \cdot \FNormS{\matY - \matX_{\mathrm{opt}}
\matX_{\mathrm{opt}}^\transp  \matY} + 4\varepsilon^2. $$
\end{theorem}

\subsection{Discussion}
Several remarks are necessary regarding our main theorem. 
First of all, notice that the notion of approximation in clustering quality is
with respect to the objective value of $k$-means; indeed, this is often the
case in approximation algorithms for $k$-means
clustering~\cite{cohen2014dimensionality}. 
We acknowledge that it would have been better to obtain results on how well
$\matX_{\tilde{\gamma}}$ approximates $\matX_{opt}$ directly, for example, via
bounding the error $\FNormS{ \matX_{\tilde{\gamma}} - \matX_{opt}}$; however,
such results are notoriously difficult to obtain since this is a combinatorial
objective. 

Next, let us take a more careful look at the effect of the parameter $\gamma_k$.
It suffices to discuss this effect for two cases: 
1) $\gamma_k = 1;$ and  
2) $\gamma_k > 1$.
To this end, we need to use a relation between the eigenvalues of $\tilde\matW$
and $\tilde\matL = \matD^{-\frac12} \matL \matD^{-\frac12}$ (see proof of
theorem for explanation): 
for all $i=1,2,\dots,n,$ the relation is: $$ \sigma_i(\tilde\matW) = 1 -
\sigma_{n-i+1}(\tilde\matL).$$

\subsubsection{$\gamma_k = 1$ }
First, we argue that the case $\gamma_k = 1$ is not interesting from a spectral
clustering perspective and hence, we can safely assume that this will not occur
in practical scenarios. 
$\gamma_k $ is the multiplicative eigen-gap between the $k$th and the $(k+1)th$
eigenvalue of $\tilde\matW$:
$$
\gamma_k =  \frac{ \sigma_k\left(\tilde{\matW}\right)}{\sigma_{k+1} \left(
\tilde{\matW} \right)} = \frac{1 - \lambda_{n-k+1}(\tilde\matL)}{ 1-
\lambda_{n-k}(\matL)}.
$$
The folklore belief~(see end of Section 4.3 in~\cite{lee2012multi}) in spectral
clustering says that a good $k$ to select in order to cluster the data via
spectral clustering is when
$$
\lambda_{n-k+1}(\tilde\matL) \ll \lambda_{n-k}(\tilde\matL).
$$
In this case, the spectral gap is sufficiently large and  $\gamma_k$ will not
be close to one; hence, a small number of power iterations $p$ will be enough
to obtain accurate results. 
To summarize, if $\gamma_k = 1,$ it does not make sense to perform spectral
clustering with $k$ clusters and the user should look for a $k'>k$ such
that the gap in the spectrum, $\gamma_{k'}$ is not approaching and at the very
least is strictly larger than $1$. 

This gap assumption is not surprising from a linear algebraic perspective as
well. 
It is well known that in order the power iteration to succeed to find the
eigenvectors of any symmetric matrix, the multiplicative eigen-gap in the
spectrum should be sufficiently large, as otherwise the power method will
not be able to distinguish the $k$th eigenvector from the $(k+1)$th
eigenvector. 
For a more detailed discussion of this we refer the reader to~Thm 8.2.4
in~\cite{golub2012matrix}.

\subsubsection{$\gamma_k > 1$}

We remark that the graph we construct to pursue spectral clustering is a
complete graph, hence it is connected; a basic fact in spectral graph
theory~(see, for example,~\cite{lee2012multi}) says that the smallest
eigenvalue of the Laplacian matrix $\matL$ of the graph equals to zero
($\lambda_n(\matL)=0$) if and only if the graph is connected. 
An extension of this fact says that the number of disconnected components in
the graph equals the multiplicity of the eigenvalue zero in the Laplacian
matrix $\matL$. 
When this happens, we have
$$\sigma_n(\matL) = \sigma_{n-1}(\matL) = ... = \sigma_{n-k+1}(\matL) = 0,$$
and, correspondingly, 
$$\sigma_n(\tilde\matL) = \sigma_{n-1}(\tilde\matL) = ... =
\sigma_{n-k+1}(\tilde\matL) = 0.$$

What happens, however, when the $k$ smallest eigenvalues of the normalized
Laplacian are not zero but close to zero? 
Cheeger's inequality and extensions in~\cite{lee2012multi} indicate that as
those eigenvalues approaching zero, the graph is approaching a graph with $k$
disconnected components, hence clustering such graphs should be ``easy'', that
is, the number of power iterations $p$ should be small. 
We formally argue about this statement below. 
First, we state a version of the Cheeger's inequality that explains the
situation for the $k=2$ case (we omit the details of the high order extensions 
in~~\cite{lee2012multi}). 
The facts below can be found in Section 1.2 in~\cite{bandeira2013cheeger}). 
Recall also the graph partitioning problem in Definition~\ref{def1}. 
Let the minimum value for $\Ncut(A,B)$ be obtained from a partition into two
sets $A_{opt}$ and $B_{opt},$ then, 
implies
$$ 
\frac{1}{2}   \lambda_{n-1}(\tilde\matL) \le \Ncut(A_{opt}, B_{opt}) \le 2
\sqrt{2 \lambda_{n-1}(\tilde\matL) }.
$$
Hence, if $\lambda_{n-1}(\tilde\matL) \rightarrow 0,$ then $\Ncut(A_{opt},
B_{opt}) \rightarrow 0,$ which makes the clustering problem ``easy'', hence
amenable to a small number of power iterations. 

\begin{figure}
\centering
\includegraphics[width=.475\textwidth]{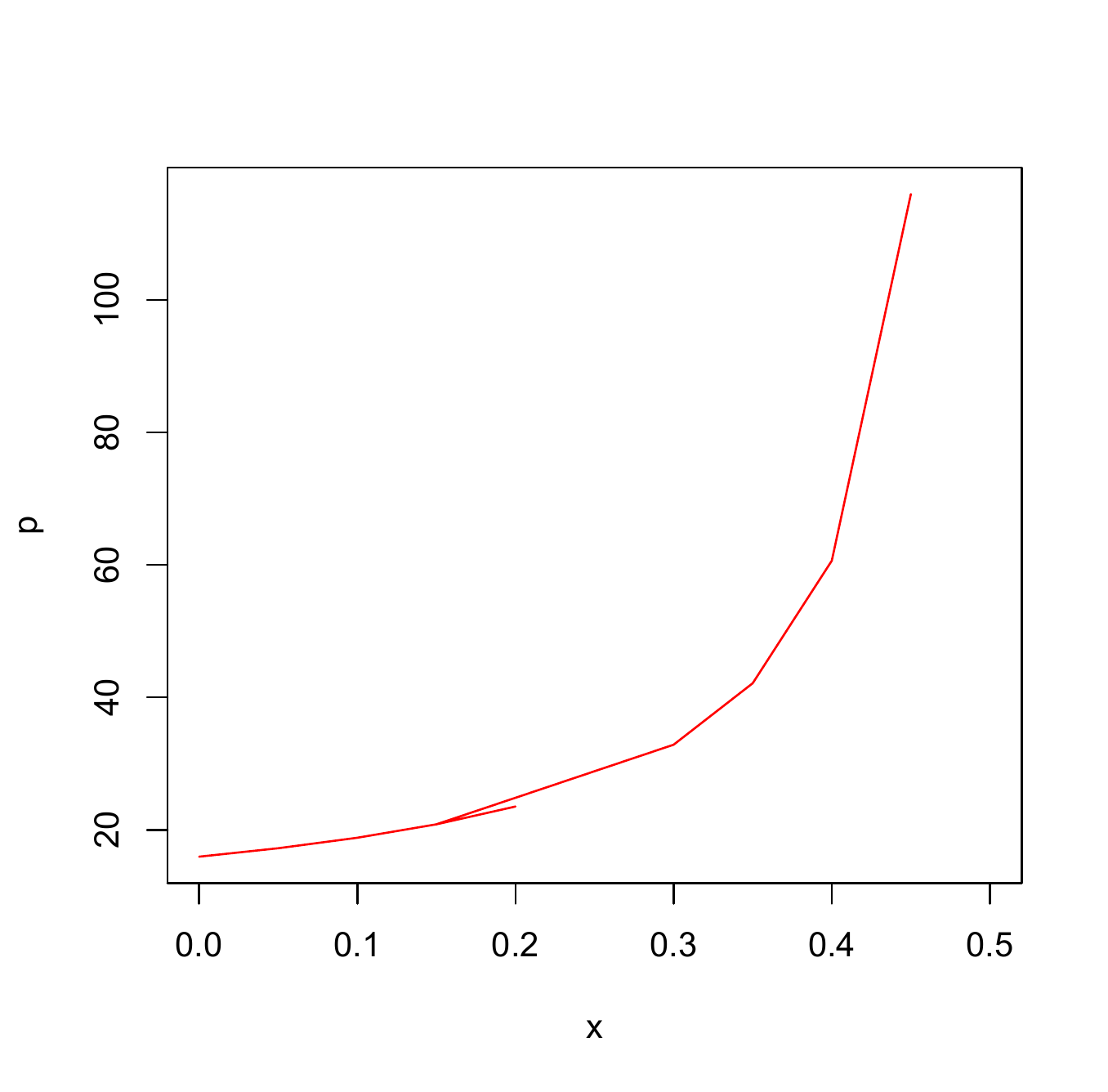} 
\caption{
Number of power iterations $p$ vs the eigenvalue $x:
=\sigma_{n-k+1}(\tilde\matL)$ of the normalized Laplacian. 
}
\label{fig:fx}
\end{figure}

Now, we formally derive the relation of $p$ as a function of
$\sigma_{n-k+1}(\tilde\matL)$.
Towards this end, fix all values ($n, \varepsilon, \delta,
\sigma_{n-k}(\tilde\matL)$) to constants, e.g., $n=10^3,$ $\varepsilon =
10^{-3},$ $\delta = 10^{-2},$ and $ \sigma_{n-k}(\tilde\matL) = 1/2$. 
Also, $x =  \sigma_{n-k+1}(\tilde\matL)$. 
Then, 
$$ 
p:=f(x) = \frac{1}{2} \frac{\ln( 4 \cdot 10^9 )}{ \ln(2 -  2\cdot x) }.
$$
Simple calculus arguments show that as $0\le x < 1/2$ (this range for $x$ is
required to make sure that $\gamma_k > 1$) increases, then $f(x)$ also
increases, which confirms the expectation that for an eigenvalue
$x:=\sigma_{n-k+1}(\tilde\matL)$ approaching to $0,$ the number of power
iterations is approaching zero as well. 
We plot $f(x)$ in Figure~\ref{fig:fx}.
The number of power iterations is always small since the dependence of $p$ on
$\sigma_{n-k+1}(\tilde\matL)$ is logarithmic. 

\subsubsection{Dependence on $\varepsilon$}
Finally, we remark that the approximation bound in the theorem indicates that
the loss in clustering accuracy can be made arbitrarily small since the
dependence on $\varepsilon^{-1}$ is logarithmic with respect to the number of
power iterations. 
In particular, when $ \varepsilon \le \FNormS{\matY - \matX_{\mathrm{opt}}
\matX_{\mathrm{opt}}^\transp  \matY}$, we have a relative error bound:
$\FNormS{\matY - \matX_{\tilde{\gamma}} \matX_{\tilde{\gamma}}^\transp  \matY}
\le $ \eqan{
&\leq&  \left( \left(1+4\varepsilon\right) \cdot  \gamma + 4\varepsilon \right)
\FNormS{\matY - \matX_{\mathrm{opt}} \matX_{\mathrm{opt}}^\transp  \matY} \\
&\leq& (1+8\varepsilon) \cdot  \gamma \cdot \FNormS{\matY -
\matX_{\mathrm{opt}} \matX_{\mathrm{opt}}^\transp  \matY},  }
where the last relation uses $1 \le \gamma$. 
One might wonder that to achieve this relative error performance, $\varepsilon
\rightarrow 0,$ in which case it appears that $p$ is very large. 
However, this is not true since the event that $\varepsilon \rightarrow 0$ is
necessary occurs only when $\FNormS{\matY - \matX_{\mathrm{opt}}
\matX_{\mathrm{opt}}^\transp  \matY} \rightarrow 0,$ which happens when $\matY
\approx \matX_{opt}$. 
In this case, from the discussion in the previous section, we have
$\sigma_{n-k+1}(\tilde\matL) \rightarrow 0,$ because $\matY$ is ``close'' to an
indicator matrix if and only if the graph is ``close'' to having $k$
disconnected components, which itself happens if and only if 
$ \sigma_{n-k+1}(\tilde\matL) \rightarrow 0.$
It is easy now to calculate that, when this happens, $p \rightarrow 0$ because,
from standard calculus arguments, we can derive: 
$$
\lim_{x \rightarrow 0}  \frac{ \frac{1}{2} \cdot \ln\left(4 \cdot n \cdot
\frac{1}{x} \cdot \delta^{-1} \cdot \sqrt{k} \right)} {\ln\left(  \frac{1 - x}{
1-  \sigma_{n-k}(\tilde\matL) } \right) } = 0.
$$ 

\section{Proofs}

\subsection{Proof of Lemma~\ref{thm:main}}\label{sec:proof}

\subsubsection{Preliminaries}
We first introduce the notation that we use throughout the paper.
\math{\matA,\matB,\ldots} are matrices; \math{\a,\b,\ldots} are column vectors.
$\matI_{n}$ is the $n \times n$ identity matrix;  $\bm{0}_{m \times n}$ is the
$m \times n$ matrix of zeros; $\bm{1}_n$ is the $n \times 1$ vector of ones.
The Frobenius and the spectral matrix-norms are $ \FNormS{\matA} = \sum_{i,j}
\matA_{ij}^2$ and $\TNorm{\matA} = \max_{\TNorm{\x}=1}\TNorm{\matA\x}$,
respectively.
The thin (compact) SVD of $\matA \in \R^{m \times n}$ of rank $\rho$ is
\begin{eqnarray}\label{eqn:svd}
\label{svdA} \matA
         = \underbrace{\left(\begin{array}{cc}
             \matU_{k} & \matU_{\rho-k}
          \end{array}
    \right)}_{\matU_{\matA} \in \R^{m \times \rho}}
    \underbrace{\left(\begin{array}{cc}
             \matSig_{k} & \bf{0}\\
             \bf{0} & \matSig_{\rho - k}
          \end{array}
    \right)}_{\matSig_\matA \in \R^{\rho \times \rho}}
    \underbrace{\left(\begin{array}{c}
             \matV_{k}^\transp\\
             \matV_{\rho-k}^\transp
          \end{array}
    \right)}_{\matV_\matA^\transp \in \R^{\rho \times n}},
\end{eqnarray}
with singular values \math{\sigma_1\left(\matA\right)\ge\ldots\geq \sigma_k\left(\matA\right)\geq\sigma_{k+1}\left(\matA\right)\ge\ldots\ge\sigma_\rho\left(\matA\right) > 0}.
The matrices $\matU_k \in \R^{m \times k}$ and $\matU_{\rho-k} \in \R^{m \times
(\rho-k)}$ contain the left singular vectors of~$\matA$; and, similarly, the
matrices $\matV_k \in \R^{n \times k}$ and $\matV_{\rho-k} \in \R^{n \times
(\rho-k)}$ contain the right singular vectors.  $\matSig_k \in \R^{k \times k}$
and $\matSig_{\rho-k} \in \R^{(\rho-k) \times (\rho-k)}$ contain the singular
values of~$\matA$.
Also, $\matA^\dagger = \matV_\matA \matSig_\matA^{-1} \matU_\matA^\transp$ denotes the pseudo-inverse of $\matA$.
For a symmetric positive definite matrix (SPSD) $\matA=\matB\matB^\transp$, $\lambda_{i}\left(\matA\right) = \sigma_{i}^2\left(\matB\right)$ denotes the $i$-th eigenvalue of $\matA$.

\subsubsection{Intermediate lemmas}
To prove  Lemma~\ref{thm:main}, we need the following simple fact. 
\begin{lemma}\label{thm:proof}
For any $\matY \in \R^{n \times k}$ and $\tilde{\matY} \in \R^{n \times k}$ with 
$\matY^\transp \matY = \tilde{\matY}^\transp \tilde{\matY} = \matI_k$:
$$
\FNormS{\matY \matY^\transp - \tilde{\matY}\tilde{\matY}^\transp}   \leq 2k \TNormS{ \matY \matY^\transp - \tilde{\matY} \tilde{\matY}^\transp }.
$$
\end{lemma}
\begin{proof}
This is because 
$\FNormS{\matX} \le \rank(\matX) \TNormS{\matX},$ for any $\matX$,
and the fact that 
$\rank(\matY  \matY^\transp - \tilde{\matY} \tilde{\matY}^\transp) \le 2k.$ 
To justify this, notice
that $\matY  \matY^\transp - \tilde{\matY} \tilde{\matY}^\transp$ can be written as the product of two matrices each with rank at most $2k$:
$$ \matY  \matY^\transp - \tilde{\matY} \tilde{\matY}^\transp =  \left(\begin{array}{cc}
             \matY & \tilde{\matY}
          \end{array}
    \right) \left(\begin{array}{c}
             \matY^\transp \\
             -\tilde{\matY}^\transp
          \end{array}
    \right).
$$
\end{proof}
We also need the following result, which appeared as 
Lemma 7 in~\cite{BM13}. 
\begin{lemma}\label{cor1}
For any matrix $\matA \in \R^{m \times n}$ with rank at least $k,$ let $p \geq 0$ be an integer and draw $\matS \in \R^{n \times k},$ a matrix of i.i.d. standard Gaussian random variables.
Fix $\delta \in (0,1)$ and $\varepsilon \in (0,1)$. Let
$\matOmega_1 = (\matA\matA^\transp)^p\matA \matS,$
and
$\matOmega_2 = \matA_k.$
If 
$$p \geq \ln\left( 4 n \varepsilon^{-1} \delta^{-1}\right) \ln^{-1}\left(\frac{\sigma_k\left(\matA\right)}{\sigma_{k+1}\left(\matA\right)}\right), $$
then w.p. $1- e^{-2n} - 2.35\delta$,
$
\| \matOmega_1 \matOmega_1^\dagger -  \matOmega_2\matOmega_2^\dagger \|_2^2 \le \varepsilon^2. 
$
\end{lemma}

\subsubsection{Concluding the proof of Lemma~\ref{thm:main}}
Applying Lemma~\ref{cor1} with $\matA = \tilde{\matW}$ and
$$ 
p \geq  \ln\left( 4 n \varepsilon^{-1} \delta^{-1} \right) \ln^{-1}\left(\frac{\sigma_k\left(\tilde{\matW}\right)}{\sigma_{k+1}\left(\tilde{\matW}\right)}\right),
$$
gives
$$
\FNormS{\matY \matY^\transp - \tilde{\matY}\tilde{\matY}^\transp} \le \varepsilon^2 .
$$
Rescaling $\varepsilon' = \varepsilon/\sqrt{k},$ i.e., choosing $p$ as
$$
p \geq  \ln\left( 4 n \varepsilon^{-1} \delta^{-1} \sqrt{k}  \right) \ln^{-1}\left(\frac{\sigma_k\left(\tilde{\matW}\right)}{\sigma_{k+1}\left(\tilde{\matW}\right)}\right),
$$
gives
$$
\FNormS{\matY \matY^\transp - \tilde{\matY}\tilde{\matY}^\transp} \le \varepsilon^2 / k.
$$
Combining this bound with Lemma~\ref{thm:proof} we obtain:
$$
\FNormS{\matY\matY^\transp - \tilde{\matY}\tilde\matY^\transp}  \leq 2k \TNormS{ \matY \matY^\transp - \tilde{\matY} \tilde{\matY}^\transp } \le
2k \cdot (\varepsilon^2 / k) = \varepsilon^2,
$$
as advertised. 

\subsection{Proof of Theorem~\ref{thm:MAIN}}\label{sec:proofmain}
Let 
$$\matY\matY^\transp = \tilde\matY \tilde\matY^\transp + \matE,$$ where $\matE$ is an $n \times n$ matrix 
with $\FNorm{\matE} \le \varepsilon $  (this follows after taking square root on both sides in the inequality of Lemma~\ref{thm:main}). 
Next, we manipulate the term $\FNorm{\matY - \matX_{\tilde{\gamma}} \matX_{\tilde{\gamma}}^\transp  \matY}$ as follows
$$ \FNorm{\matY - \matX_{\tilde{\gamma}} \matX_{\tilde{\gamma}}^\transp  \matY}  = $$
\eqan{
&=& \FNorm{\matY\matY^\transp - \matX_{\tilde{\gamma}} \matX_{\tilde{\gamma}}^\transp  \matY \matY^\transp} \\
&=& \FNorm{\tilde\matY \tilde\matY^\transp + \matE - \matX_{\tilde{\gamma}} \matX_{\tilde{\gamma}}^\transp  (\tilde\matY \tilde\matY^\transp + \matE)} \\
&\le&  \FNorm{\tilde\matY \tilde\matY^\transp - \matX_{\tilde{\gamma}} \matX_{\tilde{\gamma}}^\transp \tilde\matY \tilde\matY^\transp} + \FNorm{ ( \matI_n -   \matX_{\tilde{\gamma}} \matX_{\tilde{\gamma}}^\transp )\matE  } \\
&\le&  \FNorm{\tilde\matY \tilde\matY^\transp - \matX_{\tilde{\gamma}} \matX_{\tilde{\gamma}}^\transp \tilde\matY \tilde\matY^\transp} + \FNorm{ \matE  } \\
&=&  \FNorm{\tilde\matY - \matX_{\tilde{\gamma}} \matX_{\tilde{\gamma}}^\transp \tilde\matY} + \FNorm{ \matE  } \\
&\le&  \sqrt{\gamma} \cdot \min_{\matX \in \cal{X}} \FNormS{\tilde\matY - \matX \matX^\transp \tilde\matY}  + \FNorm{ \matE  } \\
&\le&  \sqrt{\gamma} \cdot \FNorm{\tilde\matY - \matX_{opt} \matX_{opt}^\transp \tilde\matY} + \FNorm{ \matE  } \\
&=& \sqrt{\gamma} \cdot \FNorm{\tilde\matY \tilde\matY^\transp - \matX_{opt} \matX_{opt}^\transp \tilde\matY \tilde\matY^\transp} + \FNorm{ \matE  } \\
&=&  \sqrt{\gamma} \cdot \FNorm{\matY\matY^\transp - \matE - \matX_{opt} \matX_{opt}^\transp(\matY\matY^\transp - \matE)} + \FNorm{ \matE  } \\
&\le& \sqrt{\gamma} \cdot  \FNorm{\matY\matY^\transp - \matX_{opt} \matX_{opt}^\transp\matY\matY^\transp } + 2 \cdot \sqrt{\gamma} \cdot \FNorm{ \matE  } \\
&=& \sqrt{\gamma} \cdot \FNorm{\matY - \matX_{opt} \matX_{opt}^\transp\matY} + 2 \cdot \sqrt{\gamma} \cdot \FNorm{ \matE  } \\
&=&\sqrt{\gamma} \cdot \sqrt{\mathrm{F}_{\mathrm{opt}}} + 2 \cdot \sqrt{\gamma} \cdot  \FNorm{ \matE  } \\
&\le&\sqrt{\gamma} \cdot \sqrt{\mathrm{F}_{\mathrm{opt}}} + 2 \cdot  \sqrt{\gamma} \cdot \varepsilon 
}
In the above, we used the triangle inequality for the Frobenius norm, the fact that $( \matI_n -   \matX_{\tilde{\gamma}} \matX_{\tilde{\gamma}}^\transp )$ and $( \matI_n -   \matX_{opt} \matX_{opt}^\transp )$ are projection matrices~\footnote{A matrix $\matP$ is called a projection matrix if is square and $\matP^2 = \matP$.} 
(combined with the fact that for any projection matrix $\matP$ and any matrix $\matZ$: $\FNorm{\matP \matZ} \le \FNorm{\matZ}$),
the fact that for any matrix $\matQ$ with orthonormal columns and any matrix $\matX$: $\FNorm{\matX \matQ^\transp} = \FNorm{\matX}$, 
the fact that $1 \le \sqrt{\gamma}$ and the definition of a $\gamma$-approximation $k$-means algorithm. 

Overall, we have proved: 
\eqan{
 \FNorm{\matY - \matX_{\tilde{\gamma}} \matX_{\tilde{\gamma}}^\transp  \matY}  
 &\le&  
 \sqrt{\gamma} \cdot \sqrt{ \mathrm{F}_{\mathrm{opt}}} + 2 \cdot  \sqrt{\gamma} \cdot \varepsilon.
}
Taking squares on both sides in the previous inequality gives:
$$ \FNormS{\matY - \matX_{\tilde{\gamma}} \matX_{\tilde{\gamma}}^\transp  \matY}  \le $$ 
\eqan{ 
&\le&  
\gamma \cdot \mathrm{F}_{\mathrm{opt}} + 4 \cdot \varepsilon \cdot \gamma\sqrt{ \mathrm{F}_{\mathrm{opt}}} + 4 \cdot \gamma \cdot \varepsilon^2 \\
&=& \gamma \cdot \left( \mathrm{F}_{\mathrm{opt}} + 4 \cdot \varepsilon \sqrt{ \mathrm{F}_{\mathrm{opt}}} + 4 \cdot\varepsilon^2 \right) \\
}
Finally, using $\sqrt{\mathrm{F}_{\mathrm{opt}}} \le \mathrm{F}_{\mathrm{opt}}$ shows the claim in the theorem. This bound fails with probability at most $e^{-2n} + 2.35\delta + \delta_{\gamma},$ which simply follows by taking the union bound on the failure probabilities of Lemma~\ref{thm:main} and the $\gamma$-approximation $k$-means algorithm.

\paragraph{Connection to the normalized Laplacian.}
Towards this end, we need to use a relation between the eigenvalues of $\tilde\matW$ and the eigenvalues of 
$\tilde\matL = \matD^{-\frac12} \matL \matD^{-\frac12}.$ 
Recall that the Laplacian matrix of the graph is 
$\matL = \matD - \matW$
and the normalized Laplacian matrix is  
$\tilde\matL = \matI_n - \tilde\matW.$ 
From the last relation, it is easy to see that an eigenvalue of $\tilde\matL$
equals to $1$ minus some  eigenvalue of $\tilde\matW$; the ordering of the eigenvalues, however, is different. 
Specifically, for $i=1,2\dots,n,$ the relation is
$$ \lambda_i(\tilde\matW) = 1 - \lambda_{n-i+1}(\tilde\matL),$$
where the ordering is: 
$$ 
\lambda_1(\tilde\matW) \ge  \lambda_2(\tilde\matW) \ge \dots \ge  \lambda_n(\tilde\matW),
$$
and
$$ 
\lambda_1(\tilde\matL) \ge  \lambda_2(\tilde\matL) \ge \dots \ge  \lambda_n(\tilde\matL). 
$$
From this relation:
$$
\gamma_k =  \frac{ \sigma_k\left(\tilde{\matW}\right)}{\sigma_{k+1} \left(
\tilde{\matW} \right)} = \frac{1 - \lambda_{n-k+1}(\tilde\matL)}{ 1- \lambda_{n-k}(\matL)}.
$$


\section{Experiments}
\label{sec:experiments}

To conduct our experiments, we developed high-quality MATLAB versions of the 
spectral clustering algorithms.
In the remainder of this section, we refer to the clustering algorithm in~\cite{shi2000normalized} as
``exact algorithm''. 
We refer to the modified version that uses the power method as
``approximate algorithm''.
To measure clustering quality, we used normalized mutual
information~\cite{Manning_2008_IIR}: 
$$NMI(\Omega;\mathrm{C}) =
\frac{I(\Omega;\mathrm{C})}{\frac{1}{2}\left(H(\Omega)+H(\mathrm{C})\right)}.
$$
Here, 
$$
\Omega=\{\omega_1,\omega_2,..,\omega_k\}
$$ 
is the set of discovered clusters and 
$$
\mathrm{C}=\{\mathrm{c}_1,\mathrm{c}_2,..,\mathrm{c}_k\}
$$ 
is the
set of true class labels.
Also,
$$I(\Omega;\mathrm{C}) = \sum_k \sum_j P(\omega_k \cap \mathrm{c}_j)\log
\frac{P(\omega_k \cap \mathrm{c}_j)}{P(\omega_k)P(\mathrm{c}_j)}
$$ is the mutual
information between $\Omega$ and $\mathrm{C}$, 
$$H(\Omega)=-\sum_k
P(\omega_k)\log P(\omega_k)
$$ 
and 
$$
H(C)=-\sum_k P(\mathrm{c}_k)\log
P(\mathrm{c}_k)
$$
are the entropies of $\Omega$ and $\mathrm{C}$, where $P$ is 
the probability that is estimated using maximum-likelihood.
NMI is a value in $[0,1]$, where values closer to 1 represent better 
clusterings.
The exact algorithm uses MATLAB's \texttt{svds} function to compute the
top-$k$ singular vectors.
The approximate algorithm exploits the tall-thin structure of $\matB$
and computes $\tilde{\matY}$ using MATLAB's
\texttt{svd} function~\footnote{\texttt{[U,S,V] = svd(B'*B);
tildeY=B*V*S\textasciicircum(-.5);}}.
The approximate algorithm uses MATLAB's \texttt{normrnd} function to 
generate the random Gaussian matrix $\matS$.
We used MATLAB's
\texttt{kmeans} function with the options \texttt{`EmptyAction', `singleton',
`MaxIter', 100, `Replicates', 10}.
All our experiments were run using MATLAB 8.1.0.604 (R2013a) on a 1.4 GHz Intel
Core i5 dual-core processor running OS X 10.9.5 with 8GB 1600 MHz DDR3 RAM.
Finally, all reported running times are for computing $\matY$ for the exact
algorithm and $\tilde{\matY}$  for the approximate algorithm (given $\tilde\matW$). 
%

%
\begin{table}[t]
\begin{center}
\begin{tabular}{|c|c|c|c|c|}
\hline
Name & $n$ & $d$ & \#nnz & $\#classes$ \\
\hline
SatImage & 4435 & 36 & 158048 & 6 \\
\hline
Segment & 2310 & 19 & 41469 & 7 \\
\hline
Vehicle & 846 & 18 & 14927 & 4 \\
\hline
Vowel & 528 & 10 & 5280 & 11 \\
\hline
\end{tabular}
\caption{
The libSVM multi-class classification datasets~\cite{CC01a} used for our
spectral clustering experiments.
}
\label{tbl:clustering_datasets}
\end{center}
\end{table}
%


\subsection{Spectral Clustering Accuracy}\label{sec:exp1}
We ran our experiments on four multi-class datasets from the libSVMTools
webpage (Table~\ref{tbl:clustering_datasets}).
To compute $\matW$, we use the heat kernel:$\matW_{ij} =
e^{-(\|\x_i-\x_j\|^2)/\sigma_{ij}},$ where $\x_i \in \R^{d}$ and $\x_j \in
\R^{d}$ are the data points and $\sigma_{ij}$ is a tuning parameter;
$\sigma_{ij}$ is determined using the self-tuning method described
in~\cite{MP04}.
That is, for each data point $i$, $\x_i$ is computed to be the Euclidean
distance of the $\ell^{th}$ furthest neighbor from $i$; then $\sigma_{ij}$ is set
to be $\x_i \x_j$ for every $(i,j)$; in our experiments, we report the results
for $\ell=7$.
\footnotesize
\begin{table*}[t]
\begin{center}
\begin{tabular}{|c|c|c|c|c|c|c|c|}
\hline
     & \multicolumn{2}{|c}{Exact}
     & \multicolumn{5}{|c|}{Approximate}\\\cline{2-8}
Name & & & \multicolumn{2}{|c|}{p=2}
         & \multicolumn{3}{|c|}{Best under exact time} \\\cline{4-8}
     & NMI & time (secs) & NMI & time (secs) & NMI & time (secs) & p \\
\hline
SatImage & 0.5905 & 1.270 & 0.5713 & 0.310 & 0.6007 & 0.690 & 6  \\
\hline
Segment  & 0.7007 & 1.185 & 0.2240 & 0.126 & 0.5305 & 0.530 & 10 \\
\hline
Vehicle  & 0.1655 & 0.024 & 0.2191 & 0.009 & 0.2449 & 0.022 & 6  \\
\hline
Vowel    & 0.4304 & 0.016 & 0.3829 & 0.003 & 0.4307 & 0.005 & 3  \\
\hline
\end{tabular}
\caption{
Spectral Clustering results for exact and approximate algorithms on the
datasets from Table~\ref{tbl:clustering_datasets}.
For approximate algorithms, we report two sets of numbers: (1) the NMI achieved
at $p=2$ along with the time and (2) the best NMI achieved while taking less
time than the exact algorithm. 
We see that the approximate algorithm performs \textit{at least} as good as
the exact algorithm for all but the Segment dataset.
%
%
}
\label{tbl:clustering_quality}
\end{center}
\end{table*}
\normalsize
%
%
To determine the quality of the clustering obtained by the approximate
algorithm and to determine the effect of $p$ (number of power iterations) on
the clustering quality and running time, we varied $p$ from $0$ to $10$.
In columns 2 and 3, we see the results for the exact algorithm, which serve as
our baseline for quality and performance of the approximate algorithm.
In columns 4 and 5, we see the NMI with 2 power iterations ($p=2$) along
with the running time.
Immediately, we see that even at $p=2$, the Vehicle dataset achieves better
accuracy than the exact algorithm while resulting in a 2.5x speedup.
The only outlier is the Segment dataset, which achieves poor NMI at $p=2$.
In columns 6---8, we report the best NMI that was achieved by the approximate
algorithm while staying under the time taken by the exact algorithm; we also
report the value of $p$ and the time taken (in seconds) for this result.
We see that even when constrained to run in less time than the exact algorithm,
the approximate algorithm bests the NMI achieved by the exact algorithm.
For example, at $p=6$, the approximate algorithm reports NMI=0.2449 for 
Vehicle, as opposed to NMI=0.1655 achieved by the exact algorithm.
We can see that in many cases, 2 subspace iterations suffice to get good
quality clustering (Segment dataset is the exception).
%
%
%
\footnotesize
\begin{figure}
\centering
\includegraphics[width=.5\textwidth]{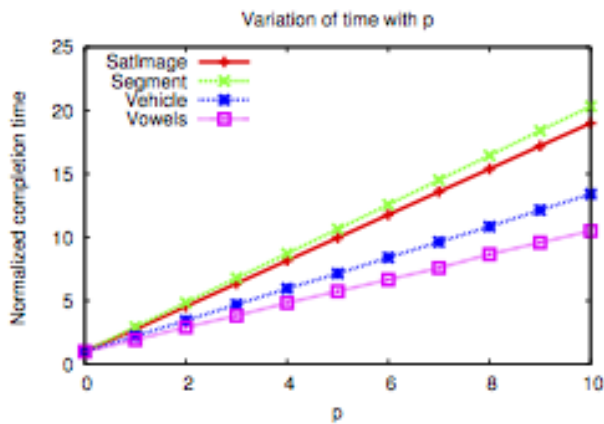} 
\caption{
Increase in running time (to compute $\tilde{\matY}$) normalized by running
time when $p=0$.
The baseline times (at p=0) are SatImage (0.0648 secs), Segment (0.0261 secs),
Vehicle (0.0027 secs), and Vowel (0.0012 secs).
}
\label{fig:clustering_timing}
\end{figure}
\normalsize

Figure~\ref{fig:clustering_timing} depicts the relation between $p$ and the
running time (to compute $\tilde{\matY}$) of the approximate algorithm.
All the times are normalized by the time taken when $p=0$ to enable reporting
numbers for all datasets on the same scale. 
As expected, as $p$ increases, we see a linear increase.

\section*{Acknowledgment} We would like to thank Yiannis Koutis and 
James Lee for useful discussions. 

\bibliographystyle{icml2015}
\bibliographystyle{plain}
\bibliography{references}

\end{document}